# Interval Probabilistic Fuzzy WordNet

Yousef Alizadeh-Q[a], Behrouz Minaei-Bidgoli[1,a], Sayyed-Ali Hossayni[a,b,c], Mohammad-R Akbarzadeh-T[b], Diego Reforgiato Recupero[d], Mohammad-Reza Rajati[f], Aldo Gangemi[d,e]

[a] *Data Mining Lab, School of Computer Engineering, Iran University of Science & Technology, Tehran, Iran*

[b] *Center of Excellence on Soft Computing and Intelligent Information Processing (SCIIP), Department of Computer Engineering, Ferdowsi University of Mashhad, Mashhad, Iran*

[c] *Agents Research Lab, TECNIO Centre EASY, ViCOROB Research Institute, Campus de Montilivi, University of Girona, Girona, Catalonia, Spain*

[d] *Semantic Technology Lab, Institute of Cognitive Sciences and Technologies (ISTC), Italian National Research Council (CNR), Italy* [e] *Laboratoire d'Informatique de Paris Nord, Université Paris 13 - Sorbonne Paris Cité - CNRS, France*

[f] *Signal and Image Processing Institute, University of Southern California, Los Angeles, CA 90089-02564*

**ABSTRACT**

WordNet lexical-database groups English words into sets of synonyms called "synsets." Synsets are utilized for several applications in the field of text-mining. However, they were also open to criticism because although, in reality, not all the members of a synset represent the meaning of that synset with the same degree, in practice, they are considered as members of the synset, identically. Thus, the fuzzy version of synsets, called fuzzy-synsets (or fuzzy word-sense classes) were proposed and studied. In this study, we discuss why (type-1) fuzzy synsets (T1 F-synsets) do not properly model the membership uncertainty, and propose an upgraded version of fuzzy synsets in which membership degrees of word-senses are represented by intervals, similar to what in Interval Type 2 Fuzzy Sets (IT2 FS) and discuss that IT2 FS theoretical framework is insufficient for analysis and design of such synsets, and propose a new concept, called Interval Probabilistic Fuzzy (IPF) sets. Then we present an algorithm for constructing the IPF synsets in any language, given a corpus and a word-sense-disambiguation system. Utilizing our algorithm and the open-American-online-corpus (OANC) and UKB word-sense-disambiguation, we constructed and published the IPF synsets of WordNet for English language.

**Keywords:** WordNet, Fuzzy WordNet, Probabilistic fuzzy WordNet, Interval probabilistic fuzzy WordNet, Interval probabilistic fuzzy synsets.

## 1 INTRODUCTION

---

[1] Corresponding author
Email addresses: alizadeh.usef@gmail.com; b_minaei@iust.ac.ir; hossayni@iran.ir; akbazar@um.ac.ir; diego.reforgiato@istc.cnr.it; rajati@usc.edu; aldo.gangemi@lipn.univ-paris13.fr;

1990, Miller et al. (Miller et al., 1990) proposed WordNet (WN), a lexical database for the English language that groups English words into synonym sets, called synsets, (also providing short definitions and usage examples) and records a number of relations among these synsets and their members. From there on, based on the WN structure, other lexical databases were also proposed for different languages (Bond & Paik, 2012)(Vossen, 1998)(Vossen, 2004) that collect synsets of their corresponding languages, as it is done in WN. We call these lexical databases under the umbrella-term WordNet-like Lexical Database (WLD). WLDs have a wide variety of applications in Natural Language Processing (Mao et al., 2018)(C. Li et al., 2019) Knowledge Engineering (C. Wang & Jiang, 2018), Ontology Engineering (Jarrar, 2019), and Ontology Embedding (Saedi et al., 2019).

In WLDs, all the members of a synset are supposed to belong to a synset with the same degree and convey the meaning of that synset at the same level. In other words, WLDs assume synsets to be crisp (non-fuzzy) sets. But this simple assumption does not always properly model the complex nature of "meaning" in natural languages. For example, let's consider the following synset of the WN: *Synset('flower.n.02'): {flower, bloom, blossom};* it contains the words that potentially (as one of their senses) stand for "reproductive organ of angiosperm plants especially one having showy or colorful parts" (the illustrative-definition of each synset is proposed in WN). Before proceeding with the mentioned issue, it is worthy to introduce the concept of a "lemma" and the concept of a "word-sense," in WLDs: Each word disregarding its various potential senses is called a *lemma*. For example, "bloom" disregarding the sense for which it can stand is considered a lemma. It is also the case for all the words of a dictionary. Moreover, a specific sense (meaning) of a lemma that is logically a member of one specific synset, is called a *word-sense*. For example, the above-mentioned sense of the lemma "bloom" is called a word-sense[2].

Usually, the lemmas (e.g. flower, bloom …) related to the word-senses of a synset (e.g. Synset('flower.n.02')), are not compatible with the meaning (definition) of the synset with the same degree. Therefore, the concept of fuzzy synsets was proposed. Since 2005, some studies have been conducted where a synset is considered a fuzzy set. In 2005, Veldall (Velldal, 2005), without using the term "fuzzy synset" (even without using the term "synset"), proposed an algorithm for creating fuzzy semantic classes[3] (i.e. synsets). In 2010, Borin and Forsberg (L Borin & Forsberg, 2010) who (to the best of our knowledge) coined the term "fuzzy synsets," viewed them from a pure linguistics point of view, and based them on "synonymy avoidance" (Hurford, 2003) concept.

In the mentioned study, Borin et al. utilized Synlex (People's synonym lexicon (Kann & Rosell, 2005) that contained synonymy[4] degree of word-pairs provided by crowdsourcing) as well as SALDO (Lars Borin, 2005)(Lars Borin & Forsberg, 2009) (a full-scale Swedish lexical-semantic resource with non-classical, associative relations among word and multiword senses[5]) by which they presented an algorithm to create the Swedish fuzzy synsets. In 2011, Gonçalo and Gomes (Gonçalo Oliveira & Gomes, 2011) were the second who looked at fuzzy synsets from a linguistics point of view. In the mentioned study, they (Gonçalo Oliveira & Gomes, 2011) applied their algorithm on Portuguese and proposed Portuguese fuzzy synsets. But the proposed fuzzy synsets are not applicable on real applications which utilize the standard and already existing WordNets. In other words, in the mentioned few studies, the synsets are either not predefined and can be determined only after running the proposed algorithm (i.e. fuzzy synsets are output of clustering (Velldal, 2005)(Gonçalo Oliveira & Gomes, 2011)), or there exist a lexical database (SALDO in (L Borin & Forsberg, 2010); yet not WN-like), but the output of the algorithm is a modified lexical database, and its synsets are not the fuzzy version of the previous synsets.

---

[2] Each lemma can have several word-senses. In other words, each lemma can be a member of more than one synset.
[3] He applied his algorithm on Norwegian language.
[4] for more information about synonymity please refer to (Osgood, 1952)

This is while, as we mentioned, WLDs have received great attention and broad applications, and those applications are based on the already existing WLDs; therefore outputting different synsets with different members means neglecting a broad variety of WLD-applications. Thus, recently, Hossayni et al. (Hossayni et al., 2016) (Hossayni et al., 2020) proposed an algorithm for fuzzification of the crisp synsets in the existing WLDs.

They propose an algorithm which assigns membership functions, for predefined synsets of any language, given a large corpus of documents of that language and a Word Sense Disambiguation (WSD)[6] as input. They, also, apply the algorithm on the English language, using the Open American National Corpus (OANC) and UKB WSD-system, by which, construct and publish the fuzzy version of WordNet.

However, in this study, we construct fuzzy-synsets whose membership degree is not a single number but is an interval. In order to construct such synsets, the theoretical framework of interval-type-2 fuzzy (IT2F) sets has to be adopted; however, we show that the existing theory of type-2 fuzzy sets is insufficient for dealing with IT2F synsets. Consequently, we propose a novel version of interval-fuzzy sets, called interval-probabilistic-fuzzy (IPF) sets, based on the probabilistic-fuzzy concept proposed in 2001 by Meghdadi and Akbarzadeh (Meghdadi & Akbarzadeh-T, 2001) and introduce IPF synsets, which contain more information than the regular fuzzy synsets. We also propose a possibility-based language-free algorithm for constructing the IPF synsets, given a language corpus and a WSD system as the input of the algorithm. Then, we apply our proposed algorithm to the English language, using the open American national corpus (OANC) and the well-known graph-based WSD system, UKB, and construct IPF synsets for English language and publish them online.

---

[5] identified by persistent formal identifiers
[6] In cognitive and computational linguistics, Word Sense Disambiguation (WSD) is an open problem belonging to ontology and natural language processing. Considering a word in a sentence, WSD identifies which of its senses is used in that sentence (for multi-sense words) (Weaver, 1955).

More in detail, we first show the missing theory for automatic production of the upgraded version (interval-valued membership degrees) of fuzzy synsets for predefined synsets, and then, propose the new mathematical concept Interval Probabilistic Fuzzy sets (IPF sets) filling the mentioned gap. Then, we present our language-free algorithm for automatically producing IPF synsets for any language, given a large corpus of documents of that language and a WSD. Finally, we apply our algorithm on English language and publish an English version of IPF synsets.

## 2    An information-lack in fuzzy synsets

Assigning a crisp membership degree to $\mu_S(x)$ (where $x$ stands for a word sense, and $S$ stands for a Synset) does not account for the uncertainty associated with the methods by which $\mu_{\text{Synset}}(\text{word-sense})$ values are computed. This yields an expectation of the possibility of occurrence of $x$ as a member of $S$; this expectation may vary according to the context which the word-sense is being used in, or the nationality/ethnicity/... of the writer/speaker. Thus, providing a crisp number to the $\mu_S(x)$ values may be effective in some contexts and ineffective in some others.

This information-lack, basically (philosophically), is quite similar to the information-lack that Mendel has proposed in Computing With Words called "intra-uncertainty" and "inter-uncertainty":

In 1999, for the first time, without using the terms "intra-uncertainty" and "inter-uncertainty", Mendel (Jerry M Mendel, 1999) presented the idea that "words mean different for different people" and therefore there exist an uncertainty considering the value of each word[7] that cannot be represented by means of T1FS whereas it can be represented by T2FS[8]. In 2003, he (J.M. Mendel, 2003) elaborated the background concepts, more, and coined the terms "intra-uncertainty" that is the uncertainty of one person (him/her self) about the value assigned to one word, and "inter-uncertainty" that is the uncertainty, related to different understandings of different people from a same word. He used IT2 FS for representation of these uncertainties. These two concepts are hot topics in Computing With Words and are continuing receiving great attention in the field (Muhuri et al., 2020)(Jerry M. Mendel, 2017)(Jerry M. Mendel, 2019).

---

[7] For those words that potentially represent a value (e.g. almost impossible, likely, somewhat law, high, tiny, sizeable, a bit, a lot, large, …)

[8] in that study, he performed a survey seeking for the numeric meaning of a word (for the words that potentially represent a measurable value) by giving questionnaires to 70 respondents asking them to assign an interval to the value they assign to each of the words(Jerry M Mendel, 1999).

Although the uncertainties, that we address, are about understandings of subjects from compatibility of meaning of word-senses with meaning (definition) of their corresponding synsets, and the uncertainties that Mendel addresses is about understanding of subjects from compatibility of meaning of a word with a range that it is addressing, the same reasoning for existence of those uncertainties in the latter ("words"-"ranges") establishes for their existence in the former ("word senses"-"synsets") as well.

However, the (common) problem of uncertainty in the membership values in fuzzy sets is normally remedied by type-2 fuzzy sets, which were introduced by Zadeh in 1975 (Zadeh, 1975). Type-2 fuzzy sets (T2 FSs) have membership degrees that are not crisp numbers but they are fuzzy sets themselves. T2 FSs are represented by a 3D membership-function (in $R^3$) in which the first axis represents the domain of the primary variable and the two other domains represent the first and the second dimension of uncertainty.

**Figure 1** graphically illustrates an example of T2 FS.

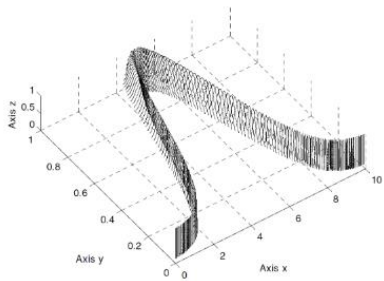 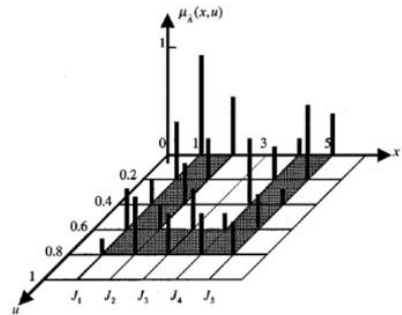

**Figure 1.** 3D illustration of a type-2 fuzzy set

**Figure 2.** Graphical illustration of an interval Type 2 Fuzzy sets. The LMF is the upper bound of the inner white rectangle and the UMF is the lower bound of the outer white rectangle (Jerry M. Mendel et al., 2016).

Sometimes, despite the insufficiency of T1FSs for modeling the uncertainty, the accessible data/information are not sufficient for constructing such a 3D fuzzy membership function. Moreover, the computations that involve T1FSs have considerable complexity. For making those computations easier and also for handling the explained cases of data insufficiency, a special case of the T2 FS is presented as an intermediate model between T1FS and T2 FS, called interval-type-2 fuzzy set (IT2 FS). In IT2 FSs, instead of a number in [0,1], a subinterval of [0,1] is assigned as the memberships. Then, membership degrees are expressed simply by their lower and upper bound.

In other words such fuzzy sets are represented by two membership functions: lower and upper (LMF and UMF). Figure 2 provides an example of IT2 FS. As it can be seen, to each value of the domain, instead of one membership degree, a range (interval) of membership degrees is assigned.

Thus, the converted IT2 FS contains the footprint of the T2 FS in its membership function (note that such membership function is a function from $\mathbb{R} \to \mathbb{R}^2$ considering that it takes an $x \in \mathbb{R}^2$ to $[a, b]$ when $a, b \in \mathbb{R}^2$). Applications of interval-valued membership degrees began from the very first appearance of T2 FSs and Zadeh's paper (Zadeh, 1975) (p. 242) and was continued being applied in other studies (Hisdal, 1981) (Schwartz, 1985) (Turksen, 1986) but establishing the bridge from T2 FSs to IT2 FS with the term "footprint" was introduced in 2000 by Liang and Mendel (Liang & Mendel, 2000) representing a bounded region that is the union of all primary membership grades of the T2 FS.

Therefore, the first idea for providing a more informative version of synsets is utilizing the IT2 FS. However, surveying the studies in the field, as far as we found, all the interval studies consider the cases which the domain of membership function is an ordered set (Algebraically speaking, "fields" (Cameron, 1998)). However, the very important point is that the members of synsets are simply "word"-senses and does not obey any order at all. Algebraically speaking, they are not even "rings" (Cameron, 1998), they are simple sets. Although as mentioned, there exist studies specifically on fuzzy sets related to "words" (Jerry M. Mendel, 2019)(Zuheros et al., 2018) that are mainly subject of the field Computing-with-Words, even those cases make a relation between words and a numeric range representing the intense of each words and words are not dealt as members of the fuzzy set.

Good news says that T2 FS is not the only 2-dimensional uncertainty. In 2001, Meghdadi and Akbarzadeh (Meghdadi & Akbarzadeh-T, 2001) proposed a new two-dimensional uncertainty, called probabilistic fuzzy set (PFS) in which membership values of elements of a set are Random Variables (RVs) (with domain [0,1]) rather than numbers. In other words, likewise T2 FSs, each member of a set can have uncountably many membership values, but to each of them "a probability density/mass value" is assigned, and not "another membership value", like what in T2 FSs (For more information about probabilistic fuzzy sets and its application, the reader is referred to (J. Li & Wang, 2017)(Gupta & Kumar, 2019).

In this study, we introduce a new version of interval-fuzzy sets analogous to IT2 with the difference that the reduced dimension, instead of being the footprint of the second fuzzy-dimension in T2 FS is the α-percent-footprint of probabilistic dimension of the PFSs (as we will define). We call this version of interval-fuzzy sets as interval possibilistic fuzzy set (IPFS) and utilize it for proposing interval-possibilistic-fuzzy synsets.

For converting a PFS to an IPFS, we have to convert RV memberships of elements of that PFS, to intervals. In other words, we convert the Probability Density/Mass Functions (PDFs/PMFs) of the corresponding RVs to simpler PDFs that their values is 1 inside a specific interval and 0, outside (they are not PMFs because they represent probability of intervals which are continuous).

For a given confidence level α (e.g. α = 0.8), for providing the above-mentioned simpler PDF of a probabilistic fuzzy member, $a$, of a PFS, $F$, we first find the interval $I^* = [\min(M_n), \max(M_n)] \subset [0,1]$ (called $\alpha$-percent-footprint) consisting of the $n$ most probable membership values of $a$ $M_n = \{m_1, m_2, \ldots, m_n\}$ for which $\sum_{x_i \in I^*} P_a(x_i) > \alpha$, ($P_a(x)$ represents the PDF/PMF of $a$) so that the above relation does not hold for the $I^* = [\min(M_{n-1}), \max(M_{n-1})]$.

Then, the (simpler) PDF of the membership RV of the probabilistic fuzzy member, $a$, of the PFS, $F$, in the IPF version of $F$, will be defined as follows

$$P_a^I(x) = \begin{cases} \dfrac{|I^* \cup I^{*C}|}{|I^*|}, & x \in I^* \\ 0, & x \notin I^* \end{cases} \quad (1)$$

where symbol I in $P_a^I(x)$ stands for interval, and C in $I^{*C}$ stands for complement (it is defined so, to assign an identical probability, inside $I^*$, 0 elsewhere, and its integral over the domain becomes 1).

Please note that the IPFS looks identical to IT2 FS (Figure 2) and the only difference is about how the interval membership grades have been provided (footprint of fuzzy uncertainty of a type-2 membership function or α-percent footprint of a probabilistic fuzzy membership function).

# 3   A possibility-based algorithm for producing Interval Probabilistic Fuzzy synsets

For applying the idea of interval-possibilistic-fuzzy sets on fuzzy synsets and proposing an algorithm for producing IPF synsets, we need an extra requirement for the algorithm input, in comparison with the input of our algorithm for fuzzy-synsets. In this algorithm, we need a corpus with text-documents from a number of ($n$) different categories. Determining the input, the algorithm is the following:

1- Applying the same process as what we perform for fuzzy-synsets on each of the categories separately. Therefore, we achieve $n$ different membership degrees for each word-sense (their accuracy was proved).

2- For each word-sense $\omega$, considering its frequency in each of the $n$ corpus-categories, and normalizing the frequencies to obtain the probability of occurrence of $\omega$ in each category, given that $\omega$ belongs to the corresponding synset.

3- Now, for each word-sense $\omega$, there exist $n$ <membership degree, probability> pairs. In other words, for each word-sense $\omega$, we have its probabilistic fuzzy membership function. Therefore, for each synset, its PFS is obtained, up to this stage of algorithm.

4- Converting the synsets' PFSs to IPFS.

The above mentioned algorithm has four phases: (1) Applying the fuzzy-synsets algorithm (introduced in the previous section) on each of the 'n' categories of the corpus and reaching to 'n' different list of fuzzy synsets. (2) Independently, computing the probability of each word-sense to be appeared in each of the 'n' categories. (3) Combining the two previous steps to reach the membership-probability value of each word-sense. (4) Computing the IPF membership values of each word-sense, based on the newly introduced concept of IPF synset. In the following you can see the pseudocode related to the mentioned algorithm.

```
For k = 1 to n //n is the number of categories
 WSF[k][][] = WSD(corpus[k]);
For i = 1 to total number of synsets
 synSize = numberOfWordSenses(synset[i]);
//"For" loop is for filling the probability mass value (PMV) matrix, containing the probability of occurring
a word-sense among other word-senses of the same synset (synset 'i').
 For k = 1 to n
```

```
  totalOccurrenceOfSynsetInK = 0;
  For j = 1 to synSize
   totalOccurrenceOfSynsetInK += WSF[k][i][j];
  For j = 1 to synSize
   If (totalOccurrenceOfSynsetInK != 0)
    PMV[i][k][j] = WSF[k][i][j] / totalOccurrenceOfSynsetInK;
   Else
    PMV[i][k][j] = NaN;
 For j = 1 to synSize
```
 {*//The block is for filling the Word Sense Probability (WSP) matrix representing the probability of occurring a word-sense in a category ('k') among the other categories.*
```
  totalOccurrenceOfWordSenseInCorpus = 0;
  For k = 1 to n
   totalOccurrenceOfWordSenseInCorpus += WSF[k][i][j];
  For k = 1 to n
   If(totalOccurrenceOfWordSenseInCorpus != 0)
    WSP[i][j][k] = <WSF[k][i][j] / totalOccurrenceOfWordSenseInCorpus,
    k>;
   Else
    WSP[i][j][k] = <NaN, k>;
```
 }{*//the block finds the largest indexes 'k's as many as the summation of WSP[i][j][k] becomes larger than α.*
```
  int[][] sorted = sort(WSP[i][j],1st);
```
 *//sorts from largest to smallest. Sorted is a 2D aray because WSP[i][j][k] is an ordered pair.*
```
  summation = 0; indexes = new int[n];
  For k = 1 to n
   summation += sorted[k][1];
    if(summation ≤ α) indexes[k]=sorted[k][2];
   Else
    indexes = indexes[1..k-1];
    break;
  sortedIndexes[i][j] = indexes;
```
 *//sortedIndexes is a 3D matrix*}
```
For i = 1 to total number of synsets
 synSize = numberOfWordSenses(synset[i]);
```

```
For j = 1 to synSize
 lowMembership1983[i][j] = +∞; upMembership1983[i][j]=-∞;
 lowMembership1993[i][j] = +∞; upMembership1993[i][j]=-∞;
```
*//"For" loop for providing possibility (membership-degree) of word-senses for each category*
```
 For k = 1 to n
  possibility3D1983[i][k][j] = 0;
  possibility3D1993[i][k][j] = 0;
  pIKj = PMV[i][k][j];
  for m = 1 to synSize
  pIKm = PMV[i][k][m];
  possibility3D1983[i][k][j] +=
   min(pIKj , pIKm);
  possibility3D1993[i][k][j] += piecewise(pIKm <= pIKj, pIKm, 0);
```
*//"For" loop for finding the foot-print of probabilistic fuzzy sets.*
```
 For k in sortedIndexes[i][j]
  lowMembership1983[i][j] =
   min(lowMembership1983[i][j],
    possibility3D1983 [i][k][j]);
  lowMembership1993[i][j] =
   min(lowMembership1993[i][j],
    possibility3D1993[i][k][j]);
  upMembership1983[i][j] =
   max(upMembership1983[i][j],
    possibility3D1983[i][k][j]);
  upMembership1993[i][j] =
   max(upMembership1993[i][j],
    possibility3D1993[i][k][j]);
```

This procedure yields to the accurate IPF membership values given that (1) Corpus is large enough (for providing the accurate probability values, as a basis for membership functions in phase 1). (2) Corpus has various enough categories to provide trustable values in phase 2). (3) WSD algorithm works precisely so that the recognized word senses will be trustable (for both phases 1 and 2).

Therefore, for applying this algorithm on English language, likewise for the previous algorithm, we used OANC (Fillmore et al., 1998), comprising almost 16.6 million words (Fillmore et al.,

1998)(de Melo et al., 2012), comprising almost 16.6 million words, in 8 different categories[9] (approximately 2.1 million words in each category) as corpus input of the algorithm, and the well-known graph-based Word Sense Disambiguation algorithm, UKB[10], and publish the entire list of English IPF synsets online[11].

It is worthy to remind that both of the proposed algorithms for producing IPF synsets is language-free and the interested researcher can apply them on his favorite language.

It is also necessary mentioning that, in the various branches of science that WN or other WordNets are utilized (mentioned in the introduction section), our published IPF synsets (and the potential output of our algorithms for other languages) can be utilized; but our synsets contain much more information than their crisp version and provide the potency of being utilized for uncertain approaches to the related branches.

## 4  Conclusions and Future work

In this study, we propose an upgraded version of fuzzy synsets in which the membership degrees are interval, similar to what in Interval Type 2 Fuzzy Sets (IT2 FSs); however, we showed that the IT2 FSs theoretical framework is insufficient for constructing such synsets, and consequently we proposed the novel concept of Interval Probabilistic Fuzzy Sets (IP FSs) for remedying this loss. Then, based on IP FSs, we proposed IPF synsets, and then proposed an algorithm for constructing them. Finally, we applied the algorithm on English language for generating IPF membership functions for WordNet word-senses.

## 5  Acknowledgements

This research was supported by the AGAUR research grant 2013 DI 012. The work was also supported, in part, by the IN2013-48040-R (QWAVES) Nuevos métodos de automatización de la búsqueda social basados en waves de preguntas, the IPT20120482430000 (MIDPOINT) Nuevos enfoques de preservación digital con mejor gestión de costes que garantizan su sostenibilidad, and VISUAL AD, RTC-2014-2566-7 and GEPID, RTC-2014-2576-7, as well as the grup de recerca consolidat CSI-ref. 2014 SGR 1469.

---

[9] Journal, telephone, letters, fiction, non-fiction, travel guides, face-to-face, technical
[10] It is necessary to note that the UKB algorithm has been evaluated in several tasks including using WN for WSD (Loureiro & Jorge, 2020), WSD on medical domain (Yue Wang et al., 2018), improvements of Information Retrieval using WN (Yinglin Wang et al., 2020), Word Embedding[10] on WN (Z. Li et al., 2019).
[11] https://bayanbox.ir/info/9034363313227650873/IPF-synsets

## 6 Vitae

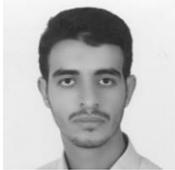 Yousef Alizadeh-Q. is a Ph.D. candidate in Artificial Intelligence (AI) at Iran University of Science and Technology (Iran). He received his Master of AI from Sharif University of Technology (SUT) (Iran) and his Bachelor of Computer Science from SUT, as well. His research interests include natural language processing, fuzzy logic, sentiment analysis, semantic word similarity, and question answering.

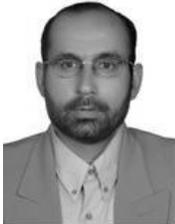 Behrouz Minaei-Bidgoli is an Associate Professor as well as the dean of School of computer engineering of Iran University of Science and Technology (IUST) and the founder of the national foundation of computer games. Ph.D. in Computer Science and Engineering (Michigan State University). Research interests: Artificial Intelligence, scientific computing, Natural Language Processing.

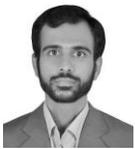 Sayyed Ali Hossayni received his Bachelor's degree in computer science from Sharif University of Technology, Iran (Sep 2004-Sep 2009), his Master's degree in computer science from Shahid Beheshti University, Iran (Sep 2010-Jan 2013), and his Ph.D. in artificial intelligence from University of Girona, Spain (2014-2018). Since Feb 2018, he is a postdoctoral researcher in data mining lab (IUST) and is mentoring more than 30 Master theses and Ph.D. dissertations, there.

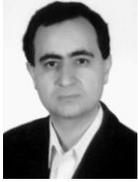Prof. Mohammad-R Akbarzadeh-T received his PhD (1998, University of New Mexico). From 1996-2002, he was also affiliated with the NASA Center for Autonomous Control Engineering at UNM. In 2006, he spent one year as a visiting scholar at UC Berkeley. In 2007, he also collaborated with Purdue University (consulting faculty). He is currently an IEEE Task force committee member, International Fuzzy Systems Association council member, and Intelligent Systems Scientific Society of Iran (ISSSI) council member. He has received several awards including the Excellent Leadership Award (2010) from IDB, Outstanding Faculty Award (2002&2008), and Outstanding Graduate Student Award (1998).

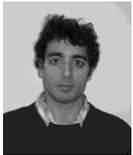Since December 2015, Diego Reforgiato Recupero has been an associate professor at the University of Cagliari, Italy. He holds a PhD from University of Naples Federico II. In 2005 he was awarded a 3 year Post Doc fellowship with the University of Maryland where he won the Computer World Horizon Award. In 2008, he won a Marie Curie International Grant to fund a 3 year Post Doc fellowship with University of Catania. He has won the "Best Researcher Award 2012", podium of the "Startup Weekend", Telecom Italia Working Capital. In March 2013 he published his first paper on SCIENCE, where he is the only author. From July 2013 to December 2015 he was a Post-Doctoral Researcher at CNR.

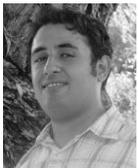Mohammad Reza Rajati the MSc. and PhD in electrical engineering and MSc in mathematical statistics respectively in 2012 and 2015, from the Department of Electrical Engineering and the Department of Mathematics, University of Southern California (USC), Los Angeles, CA, USA, where he was an Annenberg Fellow since 2009. He has published several technical papers. His research interests include computational intelligence and its applications to control systems, decision making, smart oilfield technologies, and signal processing. Dr. Rajati is already an lecturer at University of South California.

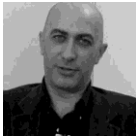Prof. Aldo Gangemi is Senior Researcher at CNR-ISTC (Rome, Italy) as well as a full professor at University of Bologna (Italy) where he coordinates the Semantic Technology research studies. He has published more than 200 refereed articles in proceedings of international conferences, journals, and books, is member of the scientific committee of Semantic Web and Applied Ontology journals, and served in dozens of program committees of international conferences and workshops.

## 7 References

Bond, F., & Paik, K. (2012). A Survey of WordNets and their Licenses. *Proceedings of the 6th Global WordNet Conference (GWC 2012)*, 64–71. http://web.mysites.ntu.edu.sg/fcbond/open/pubs/2012-gwc-wn-license.pdf

Borin, L, & Forsberg, M. (2010). From the people's synonym dictionary to fuzzy synsets-first steps. *Proc. LREC 2010 Workshop Semantic Relations. Theory and Applications.*

Borin, Lars. (2005). Mannen är faderns mormor: Svenskt associationslexikon reinkarnerat. *LexicoNordica*, *12*, 39–54.

Borin, Lars, & Forsberg, M. (2009). All in the family: A comparison of SALDO and WordNet. *Proceedings of the Nodalida 2009 Workshop on WordNets and Other Lexical Semantic Resources - between Lexical Semantics, Lexicography, Terminology and Formal Ontologies. NEALT Proceedings Series*. http://hdl.handle.net/10062/9836

Cameron, P. J. (1998). Introduction to Algebra. In *Introduction to Algebra* (Vol. 45). Oxford University Press.

de Melo, G., Baker, C. F., Ide, N., Passonneau, R. J., & Fellbaum, C. (2012). Empirical Comparisons of MASC Word Sense Annotations. *Lrec 2012 - Eighth International Conference on Language Resources and Evaluation*, 3036–3043.

Fillmore, C. J., Ide, N., Jurafsky, D., & Macleod, C. (1998). An American National Corpus: A Proposal. *The First International Language Resources and Evaluation Conference*, 965–970.

Gonçalo Oliveira, H., & Gomes, P. (2011). Automatic Discovery of Fuzzy Synsets from Dictionary Definitions. *22nd International Joint Conference on Artificial Intelligence*, 1801–1806. http://ijcai.org/papers11/Papers/IJCAI11-302.pdf

Gupta, K. K., & Kumar, S. (2019). Hesitant probabilistic fuzzy set based time series forecasting method. *Granular Computing*, *4*(4), 739–758. https://doi.org/10.1007/s41066-018-0126-1

Hisdal, E. (1981). The IF THEN ELSE statement and interval-valued fuzzy sets of higher type. *International Journal of Man-Machine Studies*, *15*(4), 385–455. https://doi.org/10.1016/S0020-7373(81)80051-X

Hossayni, S. A., Akbarzadeh-T, M. R., Reforgiato Recupero, D., Gangemi, A., & de la Rosa i Esteva, J. L. (2016). Fuzzy Synsets, and Lexicon-Based Sentiment Analysis. *Semantic Sentiment Analysis Workshop, Proceedings of 13th European Semantic Web Conference*.

Hossayni, S. A., Mohammad, A., Reforgiato Recupero, D., Gangemi, A., Del Acebo, E., & de la Rosa, J. L. (2020). An Algorithm for Fuzzification of WordNets, Supported by a Mathematical Proof. *ArXiv*.

Hurford, J. (2003). Why Synonymy is Rare: Fitness is in the Speaker. In *Ecal03* (pp. 442–451). http://www.isrl.uiuc.edu/~amag/langev/paper/hurford03ECAL.html

Jarrar, M. (2019). The Arabic ontology–an Arabic wordnet with ontologically clean content. *Applied Ontology*, 1–26.

Kann, V., & Rosell, M. (2005). Free construction of a free Swedish dictionary of synonyms. *Proc. 15th Nordic Conf. on Comp. Ling.–NODALIDA (5)*, 1–6. http://www.researchgate.net/profile/Viggo_Kann/publication/246238716_Free_construction_of_a_free_Swedish_dictionary_of_synonyms/links/0deec52ebb264a7a1c000000.pdf

Li, C., Feng, S., Zeng, Q., Ni, W., Zhao, H., & Duan, H. (2019). Mining Dynamics of Research Topics Based on the Combined LDA and WordNet. *IEEE Access*, 7, 6386–6399. https://doi.org/10.1109/ACCESS.2018.2887314

Li, J., & Wang, J. qiang. (2017). Multi-criteria Outranking Methods with Hesitant Probabilistic Fuzzy Sets. *Cognitive Computation*, 9(5), 611–625. https://doi.org/10.1007/s12559-017-9476-2

Li, Z., Yang, F., & Luo, Y. (2019). Context Embedding Based on Bi-LSTM in Semi-Supervised Biomedical Word Sense Disambiguation. *IEEE Access*, 7, 72928–72935. https://doi.org/10.1109/ACCESS.2019.2912584

Liang, Q., & Mendel, J. M. (2000). Interval type-2 fuzzy logic systems: theory and design. *IEEE Transactions on Fuzzy Systems*, 8(5), 535–550. https://doi.org/10.1109/91.873577

Loureiro, D., & Jorge, A. M. (2020). Language modelling makes sense: Propagating representations through wordNet for full-coverage word sense disambiguation. *ACL 2019 - 57th Annual Meeting of the Association for Computational Linguistics, Proceedings of the Conference*, 5682–5691. https://doi.org/10.18653/v1/p19-1569

Mao, R., Lin, C., & Guerin, F. (2018). Word embedding and wordnet based metaphor identification and interpretation. *ACL 2018 - 56th Annual Meeting of the Association for Computational Linguistics, Proceedings of the Conference (Long Papers)*, 1, 1222–1231. https://doi.org/10.18653/v1/p18-1113

Meghdadi, A. H., & Akbarzadeh-T, M.-R. (2001). Probabilistic Fuzzy Logic and Probabilistic Fuzzy Systems. *The 10th IEEE International Conference on Fuzzy Systems*, 1127–1130.

Mendel, J.M. (2003). Fuzzy sets for words: a new beginning. *The 12th IEEE International Conference on Fuzzy Systems, 2003. FUZZ '03.*, 1, 37–42. https://doi.org/10.1109/FUZZ.2003.1209334

Mendel, Jerry M. (2017). Sources of Uncertainty. *Uncertain Rule-Based Fuzzy Systems*, 245–258. https://doi.org/10.1007/978-1-4842-0184-8_2

Mendel, Jerry M. (2019). Type-2 fuzzy sets as well as computing with words. *IEEE Computational Intelligence Magazine*, 14(1), 82–95. https://doi.org/10.1109/MCI.2018.2881646

Mendel, Jerry M., Rajati, M. R., & Sussner, P. (2016). On clarifying some definitions and notations used for type-2 fuzzy sets as well as some recommended notational changes. *Information Sciences*. https://doi.org/10.1016/j.ins.2016.01.015

Mendel, Jerry M. (1999). Computing With Words, When Words Can Mean Different things to Different People. In *Proceedings of the 3rd International ICSC Symposium on Fuzzy Logic and Applications, Rochester, NY* (pp. 158–164).

Miller, G. a., Beckwith, R., Fellbaum, C., Gross, D., & Miller, K. J. (1990). Introduction to wordnet: An on-line lexical database. *International Journal of Lexicography*, 3(4), 235–244. https://doi.org/10.1093/ijl/3.4.235

Muhuri, P. K., Gupta, P. K., & Mendel, J. M. (2020). Person footprint of uncertainty-based CWW

model for power optimization in handheld devices. *IEEE Transactions on Fuzzy Systems*, *28*(3), 558–568. https://doi.org/10.1109/TFUZZ.2019.2911049

Osgood, C. E. (1952). the Nature and Measurement of Meaning. *Psychological Bulletin*, *49*(3), 227. https://doi.org/10.1037/h0021468

Saedi, C., Branco, A., António Rodrigues, J., & Silva, J. (2019). *WordNet Embeddings*. 122–131. https://doi.org/10.18653/v1/w18-3016

Schwartz, D. G. (1985). The case for an interval-based representation of linguistic truth. *Fuzzy Sets and Systems*, *17*(2), 153–165. https://doi.org/10.1016/0165-0114(85)90053-3

Turksen, I. B. (1986). Interval valued fuzzy sets based on normal forms. *Fuzzy Sets and Systems*, *20*(2), 191–210. https://doi.org/10.1016/0165-0114(86)90077-1

Velldal, E. (2005). A fuzzy clustering approach to word sense discrimination. *Proceedings of the 7th International Conference on Terminology and Knowledge Engineering*. http://heim.ifi.uio.no/~erikve/pubs/Velldal05.pdf

Vossen, P. (2004). EuroWordNet: A multilingual database of autonomous and language-specific WordNets connected via an inter-lingual-index. *International Journal of Lexicography*, *17*(2), 161–173. https://doi.org/10.1093/ijl/17.2.161

Vossen, P. (1998). Introduction to EuroWordNet. *Computers and the Humanities*, *32*, 73–89. https://doi.org/10.1023/A:1001175424222

Wang, C., & Jiang, H. (2018). Explicit utilization of general knowledge in machine reading comprehension. *Proceedings of the 57th Annual Meeting of the Association for Computational Linguistics*.

Wang, Yinglin, Wang, M., & Fujita, H. (2020). Word Sense Disambiguation: A comprehensive knowledge exploitation framework. *Knowledge-Based Systems*, *190*. https://doi.org/10.1016/j.knosys.2019.105030

Wang, Yue, Zheng, K., Xu, H., & Mei, Q. (2018). Interactive medical word sense disambiguation through informed learning. *Journal of the American Medical Informatics Association*, *25*(7), 800–808. https://doi.org/10.1093/jamia/ocy013

Weaver, W. (1955). Translation. *Machine Translation of Languages*, *14*, 15–23. http://htl.linguist.univ-paris-diderot.fr/biennale/et09/supportscours/leon/Leon.pdf

Zadeh, L. A. (1975). The concept of a linguistic variable and its application to approximate reasoning—I. *Information Sciences*, *8*(3), 199–249. https://doi.org/10.1016/0020-0255(75)90046-8

Zuheros, C., Li, C. C., Cabrerizo, F. J., Dong, Y., Herrera-Viedma, E., & Herrera, F. (2018). Computing with Words: Revisiting the Qualitative Scale. *International Journal of Uncertainty, Fuzziness and Knowlege-Based Systems*, *26*(Suppl.2), 127–143. https://doi.org/10.1142/S0218488518400147